\documentclass[conference]{IEEEtran}
\usepackage{blindtext, graphicx}
\usepackage[utf8]{inputenc}
\usepackage[english, activeacute]{babel}
\usepackage{listings}
\usepackage{amssymb}
\usepackage[breaklinks=true]{hyperref}

\begin{document}

\title{3D Human Face Reconstruction\\
with 3DMM face model from RGB image}

\author{
\IEEEauthorblockN{Zhangnan Jiang}
\IEEEauthorblockA{Tandon School of Engineering\\
New York University\\
zj2028\\
Email: zj2028@nyu.edu.com}
\and
\IEEEauthorblockN{Zichen Yang}
\IEEEauthorblockA{Tandon School of Engineering\\
New York University\\
zy2486\\
Email: zy2486@nyu.edu}}

\maketitle

\begin{abstract}
Nowadays as convolution neural networks demonstrate its powerful problem-solving ability in the area of image processing, efforts have been made to  reconstruct detailed face shapes from 2D face images or videos. However, to make the full use of CNN, a large number of labeled data is required to train the network.  Coarse morphable face model has been used to synthesize labeled data. However, it is hard for coarse morphable face models to generate photo-realistic data with detail such as wrinkles. In this project, we present a pipeline that reconstructs a human face 3D model from a single RGB image. The pipeline includes face detection, landmark detection, regression of 3DMM model parameters, and soft rendering.\\
Mentor: Zhipeng Fan (Email: zf606@nyu.edu)\\
Code Repository: https://github.com/SeVEnMY/3d-face-reconstruction\\
Code Reference: https://github.com/sicxu/Deep3DFaceRecon\_pytorch
\end{abstract}

\IEEEpeerreviewmaketitle

\section{Project Overview}
The subtasks of this project could be described as follows:
\begin{itemize}
\item \textbf{Literature Search \& Project Planning(Zhangnan Jiang, Zichen Yang):} Study related papers and approaches to get the scope of this project, complete the project plan.
\item \textbf{Data Analysis \& Preprocessing(Zichen Yang):} Acquire datasets and analyze them. Design a method preprocessing the data to make use of it in this project. (Find possible methods for data augmentation when necessary). 
\item \textbf{Implement Facial Landmarks Extraction Function(Zhangnan Jiang, Zichen Yang):} Investigate facial landmarks extraction methods from the literature and implement basic functions.
\item \textbf{Implement BFM with DL library(Zhangnan Jiang, Zichen Yang):} Implement basic functions in BFM to generate 3D face model from the shape, texture, rotation, translation, and lighting of the face.
\item \textbf{Implement Face Render function(Zhangnan Jiang, Zichen Yang):} Implement rendered to generate 2D face image from reconstructed 3D face model to calculate image-level loss.
\item \textbf{Model \& Algorithm (Loss, Optimization) Implementation(Zhangnan Jiang, Zichen Yang):} Implement the model and algorithms designed by the literature, to get a similar or even better result.
\item \textbf{Training \& Experiment(Zhangnan Jiang, Zichen Yang):} Adjusting hyper parameters (learning rate, number of epochs, etc.) to conduct various experiments, investigate results. 
\item \textbf{Further Experiments (Zhangnan Jiang, Zichen Yang):} Adjusting training methods and procedures to conduct more. 
\item \textbf{Improvement, Optimization \& Further Experiments(Zhangnan Jiang, Zichen Yang):} Based on the implementation and training results, try to find possible improvement regarding the model, algorithm or training method. More experiments will be conducted to validate the improvement.
\item \textbf{Preparing Final Presentation \& Final Report(Zhangnan Jiang, Zichen Yang):} Update from the midterm report with accomplishments, the possible improvement, and experiment results.
\end{itemize}

\section{Project Accomplishment}
\subsection{Literature Review}
We review mainly four different recent papers that focus on the 3D human face reconstruction. \\ 
Deng, Yu et al.\cite{weakly-supervised} proposed a convolution neural network that could obtain accurate 3D face reconstruction with weakly-supervised learning. A hybrid-level loss function is proposed to use both low-level information of pixel-wise color and  the pixel-wise consistency with raw image signal. Meanwhile, a novel skin color based photo metric error attention strategy is used to make sure the network is robust to occlusion and other challenging appearance variations such as beard and heavy make-up. The process of getting a 3D model from a single image could be summarized as: Firstly, estimation and landmark detection are extracted from the input image, these two are used as weak supervision signals. Res-Net is used in this paper to generate parameters that are needed to build the 3D models. The advantages of this method is that it does not require any ground-truth 3D shapes for training. \\ 
Tran, Luan and Liu, Xiaoming\cite{nonlinear} proposes an innovative framework to learn a nonlinear 3DMM model from a large set of unconstrained face images, without collecting 3D face scans. Two decoder were introduced to serve as the nonlinear 3D Morphable Model to map from the shape and texture parameters to the 3D shape and texture. Compared to the previous one, this method is also an end-to-end trainable neural network with only weak supervision. At the same time, this method designed two separating neural network decoder for shapes and textures: the multi-layer perceptron (MLP) is used for shape decoding, while convolution neural network is used for texture. The main advantage of this method is that it could be learnt jointly with 3DMM in an end-to-end fashion. \\
On the other hand, Richardson et al.\cite{detailed} introduced an end-to-end CNN framework which derives the shape in a coarse-to-fine fashion. The network is mainly constructed by two different blocks, a network that recovers the coarse facial geometry (CoarseNet), and a CNN that refines the facial features of that geometry (FineNet). Compare to the above methods, authors trained their network in two different phase: firstly training regime begins with a supervised phase, based on
synthetic images, then it is followed by an unsupervised phase that uses only unconstrained facial images. Compared to other methods we have reviewed, this methods could demonstrates more detail information such as wrinkles. The main disadvantage is it may have bad performance when the testing case is not included in the training data, such as beards, makeup. \\
As for method \cite{dense-face} from Yudong Guo et al., they proposed a CNN-based face construction method that could used for both monocular video and single RGB image. Compared to the coarse-to-fine fashion, they added  coarse-scale tracking network (Tracking CoarseNet). At the same time, they have generated photo-realistic datasets as the training set for CoarseNet and FineNet. The major advantage of this method is it could have detailed information such as wrinkles in the reconstructed 3D models. However, it is not weakly supervising network as the methods we have discussed, it require a data set to do the training in order to get a good result. 

\subsection{Data Analysis \& Pre-processing}
For pre-processing part, we followed material given by our mentor Zhipeng Fan. Firstly, we submitted an application for downloading of Basel 3D Face Model, which is widely used in 3D face reconstruction. After downloading the BFM model, we load expression basis from the model, and then transfer original BFM09 to our face model. The face model we are going to use is cropped align face landmarks which contains only 35709 vertex. However, as for the expression basis we used here, it contains 53215 vertex, therefore we select corresponding vertex to get our face model. Meanwhile, \emph{facemodelinfo.mat} from BFM is loaded for region used for computing photometric loss, region used for skin texture regularization. \\
Moreover, We run through the numpy implementation of BFM model, which gives us a basic idea of what BFM model is constructed by. We examine the process of BFM model by plot the visualization of human face key points, the reconstructed 3D face, and the model with light. \\
\begin{figure}[!h]
\centering
\includegraphics[width=0.25\textwidth]{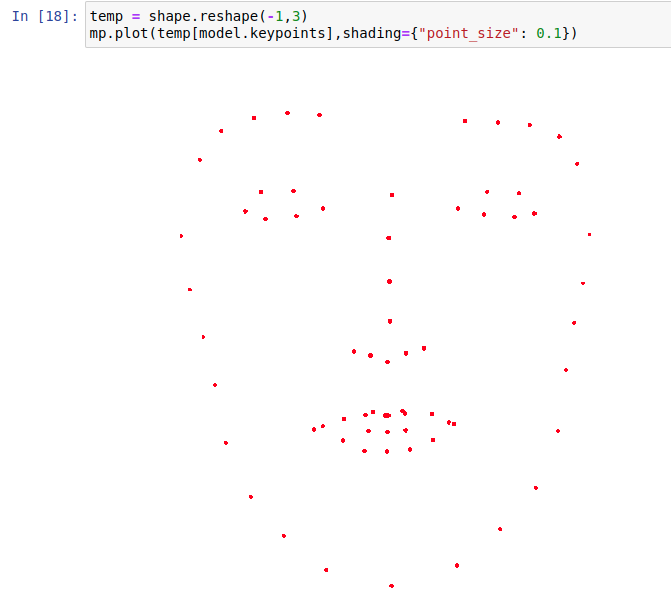}
\caption{Key points in BFM model visualization}
\label{fig1}
\end{figure}

\begin{figure}[!h]
\centering
\includegraphics[width=0.25\textwidth]{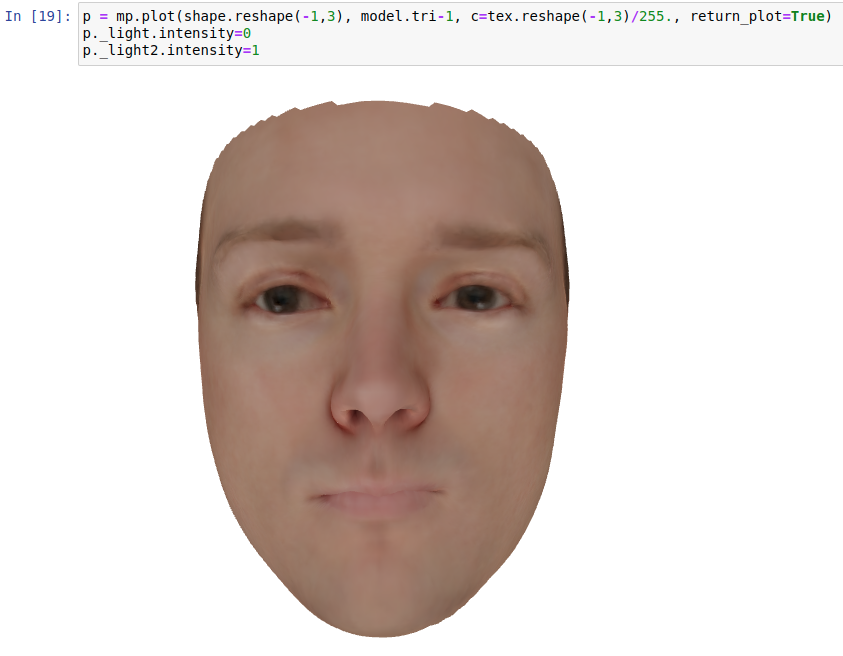}
\caption{BFM model visualization without lightning}
\label{fig2}
\end{figure}

\begin{figure}[!h]
\centering
\includegraphics[width=0.25\textwidth]{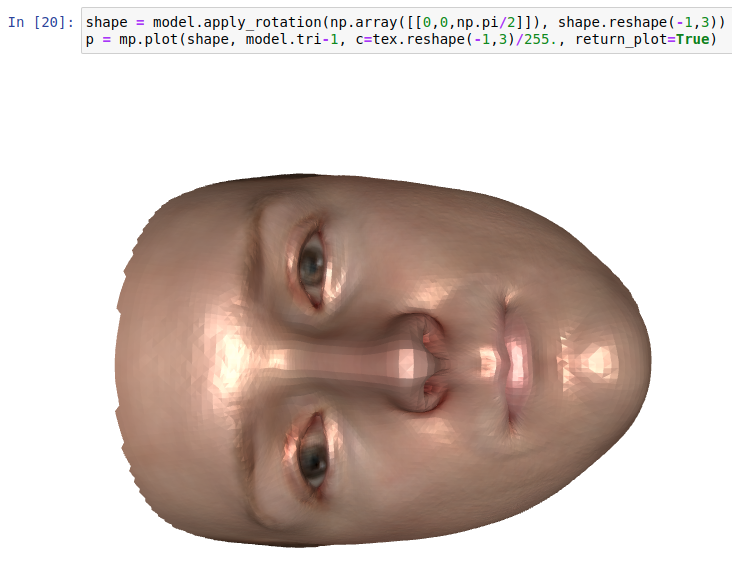}
\caption{BFM model visualization with lightning}
\label{fig3}
\end{figure}

Meanwhile, we have tried to use face alignment package in python to extract human face feature from one single RGB image. By using this package, we could extract human face feature from both the input image and the rendered image. Following is our experiment using one input image, the human face feature could be extracted properly. 

\begin{figure}[!h]
\centering
\includegraphics[width=0.25\textwidth]{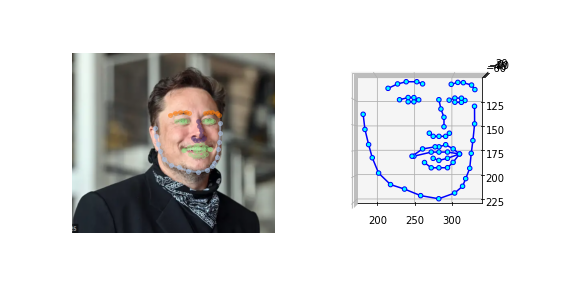}
\caption{Human face feature extracted from one single input image}
\label{fig4}
\end{figure}

After reading papers and doing researches, we decided to use Labeled Faces in the Wild\cite{lfw}(LFW). It is a public benchmark for face verification. More than 13,000Cropped and aligned human face images are provided, which are more than enough for us to train the model. Before we directly feed the image to the neural network, we use MTCNN\cite{mtcnn} to extract five most important feature points on human face, they are: left eye, right eye, nose, mouse left, mouse right. These five points would help face alignment to get more precise landmark. The combination of original image and text file contains the coordinates of these five feature points are the input of our pipeline.

\subsection{Facial Landmarks Extraction}
In this project, in order to calculate the landmark loss, we need to extract facial landmarks from input images and predicted results. Facial landmarks from input and output are compared and their difference is one of the loss. A. Bulat et al. \cite{fa} proposed a method for detecting facial landmarks using python, which makes use of the world's best-performance face alignment network. 
\begin{figure}[!h]
\centering
\includegraphics[width=0.25\textwidth]{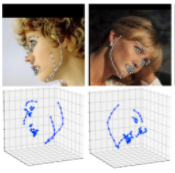}
\caption{Facial landmarks in 3D space}
\label{fig5}
\end{figure}
We make use of the 3D facial landmarks extraction method provided by this python library. The method by default detects faces in the input by S3FD face detector \cite{sfd}, the landmarks across the face detected are labeled. In our implementation, we take landmarks across the center of the whole face, both eyebrows, both eyes, nose, nostrils, lips and teeth. 

\subsection{Basel Face Model(BFM)}
There are a variety of 3D Morphable Models for 3D objects reconstruction. Pascal Paysan et al. \cite{bfm} in 2009 proposed the Basel 3D Face Model, which significantly increased the shape and texture accuracy of generating 3D face models using an optimized scanning device and less correspondence artifacts. In our project, we need to reconstruct human faces using Basel 3D Face model. The shape of the face, the color and the lighting need to be calculated and optimized by our Deep Neural Network model and passed to the Basel Face Model. In order to reconstruct a 3D Face Model by Basel approach, parameters needed are calculated as follow:
\begin{itemize}
\item \textbf{Shape:} \[s = \overline{s} + A_{id}\alpha_{id} + A_{exp}\alpha_{exp}\]
\item \textbf{Texture:} \[t = \overline{t} + A_{tex}\alpha_{tex}\]
\item \textbf{Lighting:} \[L(b_{i},n_{i}|\gamma) = b_{i}\sum_{b=1}^{B^{2}}\gamma_{k}\phi_{k}(n_{i})\]
\item \textbf{Rotation:} \[rot({x}),\space rot({y}),\space rot({z})\]
\item \textbf{Translation:} \[{T}\]
where \(\overline{s}\) is the mean shape of the face model, \(A_{id}\) is the shape identity basis, \(A_{exp}\) is the shape expression basis, \(\overline{t}\) is the mean texture of the face model, \(A_{tex}\) is the texture basis. In the lighting, \(\phi_{k}(n_{i})\) is a normalization function for the vertex \(n_{i}\), \(b_{i}\) is the albedo, all of these parameters are provided by the pre-trained Basel Face Model. In our approach, we need to calculate and optimize the identity coefficient \(\alpha_{id}\), the expression coefficient \(\alpha_{exp}\), the texture coefficient \(\alpha_{text}\) and the lighting coefficient \(\gamma\). For further optimization to get a more accurate reconstruction result, the pose coefficients including the rotation(\(R\)) and the translation(\(T\)) of the face need to be calculated and fitted into the Basel Face Model. \(rot()\) is the rotation function provided by BFM taking coefficients from three axis \(x\), \(y\) and \(z\). \(T\) is the translation coefficient. Appreciating the project mentor Zhipeng for providing the function implementation for the BFM. We made some adjustment to make it available for tensor inputs. 
\end{itemize}
\subsection{Differentiable Renderer}
In this project, we calculate 2D image loss for backpropagation and optimizing the deep neural network. When we acquire the 3D face model, we need to render the 2D face image of the 3D model, which requires the use of differentiable renderer.\\
In this project, we make use of Soft Rasterizer(SoftRas)\cite{soft} to render the 2D face image from reconstructed 3D face model generated by Basel Face model. In our implementation, we pass the 3D face model, view angle, light condition and texture type to the renderer to generate the 2D face image. After generating the 2D face image, we stitch the face image back into the original image to cover the face area, which process can help us calculating the image loss.  \\
Thanks the project mentor Zhipeng for providing the soft renderer usage example. When embeding the soft rasterizer into our project, we found several bugs related to the original code, some changes are made to the library code. 
\subsection{Model \& Algorithm Implementation}
\begin{figure*}
\centering
\includegraphics[width=0.9\textwidth]{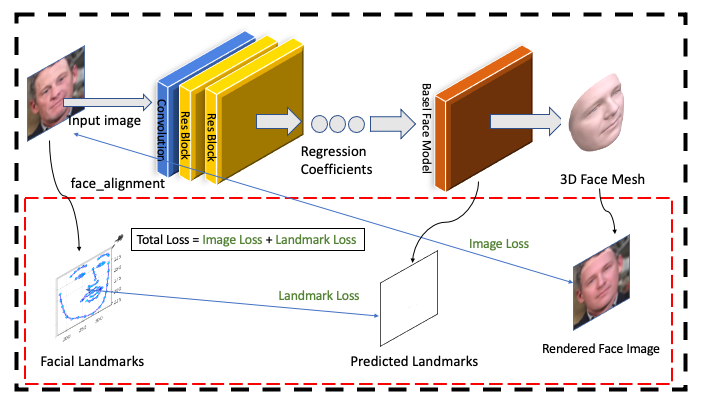}
\caption{Overview of our approach}
\label{fig6}
\end{figure*}
\subsubsection{Residual Network}
Residual Network\cite{resnet} has shown extraordinary performance on computer vision tasks including object detection, semantic segmentation etc. As the neural network going deeper, there comes up some problem including gradient vanishing and overfitting etc. RNN can efficiently solve these problems by its residual connection, in which an identity connection can propagate input information directly to next layer, concatenating with the results calculated by the convolutional layers. The method can significantly preserve the information from image input as the neural network going deeper.  In this project, we implemented a RNN based 3D face reconstruction method. This convolutional neural network extracts face features information for the learning process. \\
In this project, we use 50-layer RNN to complete the regression. The model is pretrained on imageNet dataset\cite{imagenet}. In the original task, there are 1000 features to be extracted and output. In our approach, the total number of dimensions should be 257(80 for \(\alpha_{id}\), 64 for \(\alpha_{exp}\), 80 for \(\alpha_{tex}\), 27 for \(\gamma\), 3 for \(T\), 1 for \(x\), 1 for \(y\) and 1 for \(z\)). We modify the model structure by adding fully-connected layers at the bottom of ResNew50. Once we get the losses defined, back-propagation is conducted to update and optimize the weights of the model. The neural network takes the face image as input, and the output of the network is the coefficients for BFM to reconstruct 3D face model. We regress the shape identity coefficient, shape expression coefficient, texture coefficient, light coefficient(illumination), rotation coefficients and translation coefficient.\\
To better train the network and avoid overfitting, we apply random dropout\cite{dropout} and batch normalization\cite{bn}. 
\subsubsection{Loss Calculation}
In current implementation, we apply two loss functions: image loss and landmarks loss. The preliminary structures for loss functions are as follow:
\begin{itemize}
\item \textbf{Image Loss (\(L_{p}\)):} We have a input image \(I_{in}\), after rendering the output image \(I_{out}\) from the 3DMM face model. We will make Mean Squared Error(MSE) between \(I_{in}\) and \(I_{out}\) across every pixel in \(I_{in}\) and \(I_{out}\): 
\[L_{p} = \frac{1}{N}\sum_{i=1}^{N}(I_{in_i}-I_{out_i})^2\]
where \(N\) is the number of pixels in the input or output image.
\item \textbf{Landmarks Loss (\(L_{l}\)):} We have facial landmarks of the input image \(l_{in}\), after rendering the 3DMM face model from BFM, we can extract the facial keypoints \(l_{out}\) from the generated model. We make Mean Squared Error(MSE) between \(l_{in}\) and \(l_{out}\) across every pixel in \(I_{in}\) and \(I_{out}\):
\[L_{l} = \frac{1}{n}\sum_{i=1}^{n}(l_{in_i}-l_{out_i})^2\]
where \(n\) is the number of landmarks extracted in the input or rendered image.
\item \textbf{Total Loss (\(L\)):} After calculating these two losses, we take weighted sum to generate the total loss:
\[L = w_{p}L_{p} + w_{l}L_{l}\]
where \( w_{p}\) is the weight of the image loss \(L_{p}\), \( w_{l}\) is the weight of the landmarks loss \(L_{l}\).
\end{itemize}
 \subsubsection{Model Overview}
As shown in Fig. \ref{fig6}, in one execution circle, firstly we use face\_alignment function to extract facial landmarks \(l_{in}\) from the input image \(I_{in}\). Then we take the face image as the input of the regression model. After the regression of the CNN model, we have the facial coefficients \(\alpha_{id}\), \(\alpha_{exp}\), \(\alpha_{text}\), \({x}\), \({y}\), \({z}\), \(\gamma_{k}\) and \({T}\). The coefficients are fitted into the BFM to generate the face meshes, an output image \(I_{out}\) is rendered by stitching the rendered face model back to the input image. As illustrated, Image Loss and Landmark Loss are calculated and backpropated through the model, RNN weights are updated and optimized. 

\section{Summary}
\subsection{Result}
In this project, we have conducted multiple experiments too find the best training setting. To train the model, the best setting is setting epoch number to 20 with the batch size 16, the initial learning rate is \(1e^{-3}\) using Adam optimizer\cite{adam}. The training process could be demonstrated as following: 
After the training, we saved the model with best performance: 
\begin{figure}[!h]
\centering
\includegraphics[width=0.25\textwidth]{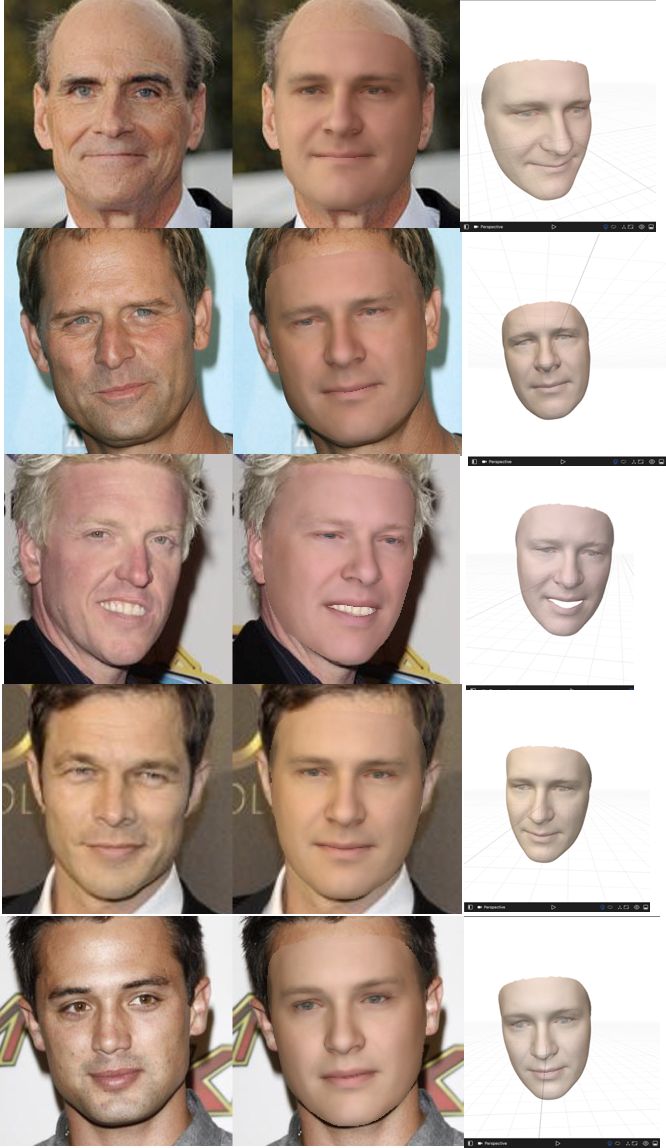}
\caption{Face Reconstruction Results}
\label{fig7}
\end{figure}

\begin{figure}[!h]
\centering
\includegraphics[width=0.25\textwidth]{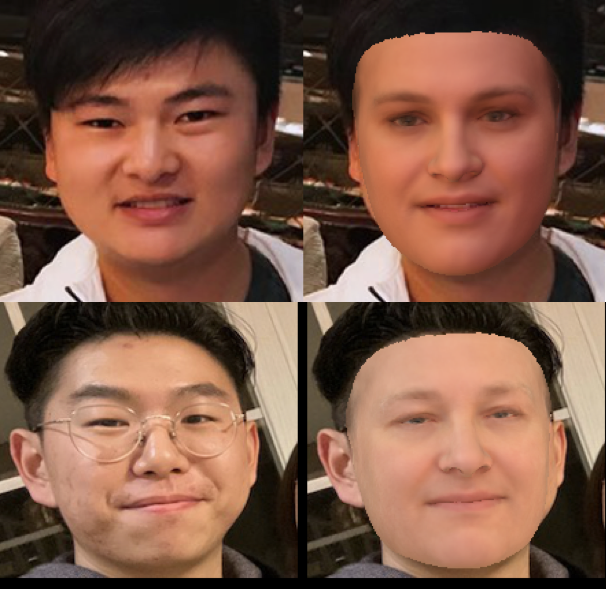}
\caption{Testing on our face images}
\label{fig8}
\end{figure}

\subsection{Reflection}
By doing this project, knowledge about neural network, computer graphics, and image feature extraction have been learned and applied. They could be summarized as following:
\subsubsection{Human Feature Extraction}
As for the part of feature extraction, we used package MTCNN\cite{mtcnn} to do the work for us. We studied the structure and how it works. MTCNN is a framework that used for both face detection and face alignment. It contains three different stages of CNN. As we discussed in the course, it also does preprocessing by building image pyramid, so that features in different scales could be detected precisely. The first stage of the MTCNN is a fully connected convolutional neural network. Bonding box regression is used in this progress. The second stage is the refine network, which outputs 2 dimension for face classification, 4 dimension for bonding box regression, and 10 for facial landmark localization. Finally, O-Net would be used to refine the result. In our project, we mainly used the 10 dimension for facial landmark. 
\subsubsection{Training Neural Network}
In this project, we used the pre-trained 50-layer RNN to help us extract features we need for 3D human face reconstruction. We have learned how to define our own loss and use the loss to do back propagation to get better parameters from the neural network. Moreover, since we are not using 3D human face data set but just human face image data set, we learned that instead of requiring ground-truth data set, we could re-define our loss function to get what we want. As in our project, the overall goal for us is to produce 3D human face from one single RGB image, instead of using the difference between ground-truth 3D object and what we have produced, we used the difference between rendered image and input image, which achieve weakly-supervised learning. 
\subsubsection{Computer Graphics}
The 3D human face object is product by BFM model. From the application of this model, we have learned the rendering pipeline. 

\subsection{Possible Improvements}

\ifCLASSOPTIONcaptionsoff
  \newpage
\fi



%\bibliographystyle{IEEEtran}

\end{document}